%% file: colm2026_conference.tex
\documentclass{article} % For LaTeX2e
\usepackage[preprint]{colm2026_conference}

\usepackage{microtype}
\usepackage{hyperref}
\usepackage{url}
\usepackage{booktabs}
\usepackage{hyperref}
\usepackage{url}
\usepackage{multirow}
\usepackage{graphicx}
\usepackage{comment}
\usepackage{subcaption}
\usepackage{booktabs}
\usepackage{algorithm}
\usepackage{algorithmic}
\usepackage[table]{xcolor}
\usepackage[dvipsnames]{xcolor}
\definecolor{lightblue}{RGB}{173,216,230}
\usepackage{lineno}

\definecolor{darkblue}{rgb}{0, 0, 0.5}
\hypersetup{colorlinks=true, citecolor=darkblue, linkcolor=darkblue, urlcolor=darkblue}

\input{math_commands.tex}

\newtheorem{assumption}{Assumption}
\newtheorem{lemma}{Lemma}
\newtheorem{remark}{Remark}
\newtheorem{theorem}{Theorem}
\newtheorem{theorem_app}{Theorem}
\newtheorem{proposition}{Proposition}
\newtheorem{proposition_app}{Proposition}
\newtheorem{definition}{Definition}

\title{Early Stopping Chain-of-thoughts in Large Language Models}

\author{Minjia Mao$^{1}$\thanks{Equal Contributions.}~~, Bowen Yin$^{2*}$, Yu Zhu$^1$ \& Xiao Fang$^1$  \\
$^1$University of Delaware, Newark, DE 19716, USA \\
$^2$Peking University, Beijing 100871, China  \\
\texttt{\{mjmao,yuzhu,xfang\}@udel.edu} , \texttt{bowenyin@stu.pku.edu.cn} \\
}
\begin{document}

\ifcolmsubmission
\linenumbers
\fi

\maketitle

\begin{abstract}

Reasoning large language models (LLMs) have demonstrated superior capacities in solving complicated problems by generating long chain-of-thoughts (CoT), but such a lengthy CoT incurs high inference costs. Previous methods on inference-stage efficient reasoning either require white-box models to monitor the reasoning process or are not reliable through direct prompting. In response, we introduce \textbf{ES-CoT}, an inference-time method that shortens CoT generation by detecting answer convergence and stopping early with almost no performance loss. When observing a linguistic marker (such as "wait") in the reasoning process, we prompt the LLM to output its current final answer, denoted as a \textit{step answer}. We then track the run length of consecutive identical step answers as a measure of answer convergence. We show both empirically and theoretically that step answers steadily converge to the final answer, and large run-length jumps reliably mark this convergence. Experiments on six reasoning datasets across three LLMs show that ES-CoT reduces the number of inference tokens by 16.08\% on average while maintaining accuracy comparable to standard CoT. 
% Implementation codes of this study are available online (hidden for peer review). 

% We then track the run length of consecutive identical step answers as a measure of answer convergence. Once the run length exhibits a sharp increase and exceeds a minimum threshold, the generation is terminated. We provide both empirical and theoretical support for this heuristic: step answers steadily converge to the final answer, and large run-length jumps reliably mark this convergence. 

% Further, ES-CoT integrates seamlessly with self-consistency prompting and remains robust across hyperparameter choices, highlighting it as a practical and effective approach for efficient reasoning. 

\end{abstract}

\section{Introduction}

Reasoning LLMs, such as OpenAI o-series models \citep{openaio1}, DeepSeek-R1 \citep{guo2025deepseek}, and QwQ \citep{qwq32b}, have achieved state-of-the-art performance on challenging tasks in mathematics, coding, and scientific reasoning \citep{li2025system}. A key driver of this progress is chain-of-thought (CoT) reasoning, which elicits intermediate reasoning steps before producing the final answer \citep{wei2022chain}. By incorporating a long thinking sequence, reasoning LLMs can plan the solution procedure, explore alternative strategies, and double-check the final result \citep{chen2024not}.

% Yet
However, longer reasoning comes at a cost. For example, recent studies reveal that LLMs frequently overthink, continue to generate redundant steps even after reaching the correct answer \citep{chen2024not}. Such verbosity inflates inference cost, aggravates memory and latency challenges, and reduces the practicality of reasoning models in real-world settings. This tension motivates the study of \textit{efficient reasoning} \citep{feng2025efficient}: how to preserve the accuracy benefits of CoT while minimizing unnecessary reasoning tokens.

% \citet{chen2024not} propose to use length preference optimization and fine-tune the model by incorporating the CoT length as an objective. However, these methods are less effective and difficult to implement in practice. 
% In response, we posit a simple yet effective hypothesis regarding the reasoning steps of a CoT sequence. Specifically, we assume that as the reasoning progresses, LLMs become more confident and move closer to their final answer \citep{prystawski2023think,qian2025demystifying}. 
In this work, we address this problem by asking: \textit{When can a reasoning trajectory be stopped without harming output quality?} 
Previous methods on inference-stage efficient reasoning either require white-box models \citep{yang2025dynamic,chen2025seal,huang2025mitigating} to monitor the reasoning process or are not reliable through direct prompting \citep{xu2025chain,ma2025reasoning,fu2025reasoning}.
In response, we introduce the concept of a \textit{step answer}, which is the model's current guess of the final answer when the model generates a reasoning linguistic marker, such as "wait" and "alternatively".
Empirical analysis across different datasets and LLMs shows a clear convergence pattern: step answers are more likely to repeat in later reasoning stages, and at some point, this repetition length makes a sharp jump, which is a signal that the LLM is committing to a stable answer (as depicted in Figure~\ref{fig:converge}). 
This observation aligns with prior findings that LLMs become increasingly confident as reasoning unfolds \citep{prystawski2023think,qian2025demystifying}. 

% need revision
% \textcolor{red}{Building on this insight, we introduce \textbf{ES-CoT (Early-Stop CoT)}, an inference-time method that halts generation once a decisive convergence signal appears. At its core is the \textit{run-jump test}: when the run length of identical step answers exhibits a statistically large leap, the reasoning is terminated, and the current step answer is returned as the final output (Figure \ref{fig:framework}).}

\begin{figure}[t]
    \centering
    \includegraphics[width=0.95\linewidth]{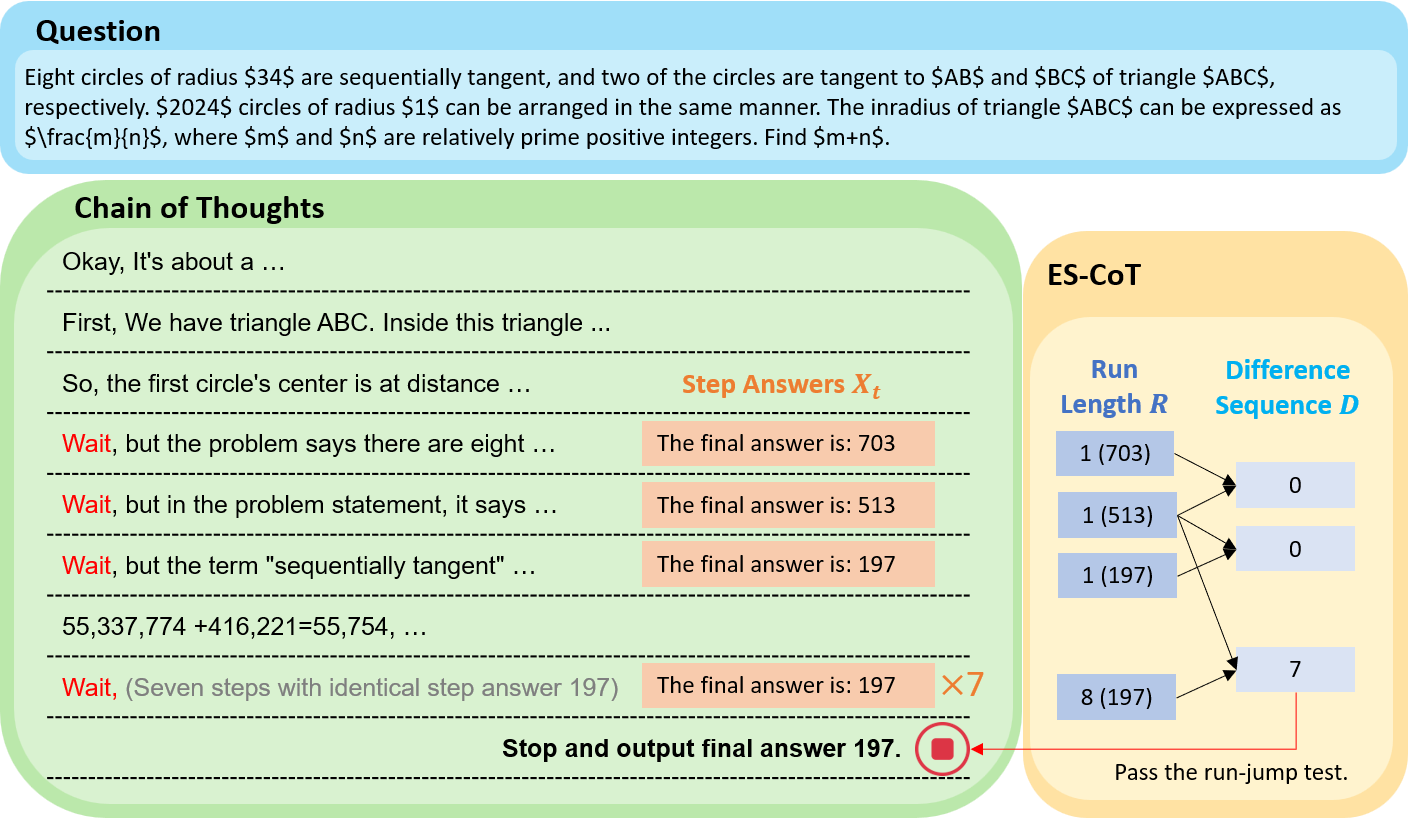}
    \caption{Framework of ES-CoT. A step answer is obtained by prompting when observing "wait".}
    \label{fig:framework}
\end{figure}

Building on this insight, we introduce \textbf{ES-CoT (Early-Stop CoT)}, an inference-time method that halts generation once a decisive convergence signal appears. Figure \ref{fig:framework} illustrates the framework of ES-CoT. As depicted, we denote a run as a sequence of consecutive steps with the same answer, and the length of the run as the number of consecutive steps. The core of ES-CoT is the \textit{run-jump test}: when the run length of identical step answers exhibits a statistically large leap, the reasoning is terminated, and the current step answer is returned as the final output, as shown on the right side of Figure \ref{fig:framework}. In short, ES-CoT offers a drop-in, supervision-free principle: stop thinking when the answer stabilizes. This makes efficient reasoning practical without additional models or retraining \citep{sui2025stop}.

% In practice, at each reasoning step, we append the prompt 'The final answer is' to ask the model for its intermediate answer (where a step is usually identified by a newline character in reasoning LLMs \citep{zheng2024processbench}). We refer to this as the step answer. Next, at each step, we collect the step answer and record consecutive identical answers. We denote a run as a sequence of consecutive steps with the same answer, and the size of the run as the number of consecutive steps.

We evaluate ES-CoT across six reasoning datasets, using three LLMs of varying scales. Experimental results demonstrate that ES-CoT consistently reduces the number of generated tokens by about 16.08\% while maintaining accuracy comparable to the original CoT prompting. Further analysis shows that ES-CoT scales robustly with different hyperparameters. 
% and integrates seamlessly with self-consistency prompting \citep{wang2023self}, yielding further gains.} 
Our contributions are threefold: 

\begin{itemize}
    \item We propose ES-CoT, an inference-time method that halts CoT when the run length of identical answers makes a statistically significant leap beyond previous runs. This design requires no extra reward model, no parallel decoding, and no retraining, as used in previous work \citep{sui2025stop}. 

    \item We show both empirically and theoretically that step answers converge toward the final answer, and that a sufficiently large run-length jump reliably marks this convergence.

    \item On six reasoning datasets with three LLMs of different scales, ES-CoT reduces token usage by about 16.08\% on average while maintaining accuracy.
\end{itemize}

% Unlike input-side prompting or model-side retraining approaches, our method requires neither additional training nor extra cost for problem-specific analysis. Instead, it adaptively performs early stopping during inference, adjusting the reasoning length across different problems and models without additional supervision. Compared to output-side methods, our approach avoids the need for parallel decoding and does not depend on an auxiliary reward model for termination.

\section{Related Work}

Our study belongs to the stream of efficient reasoning, which seeks to reduce reasoning length while preserving reasoning capabilities \citep{sui2025stop}. Prior studies in this area can be broadly categorized into three groups: input-side, model-side, and output-side efficiency. We compare previous methods with our proposed method ES-CoT in detail in Appendix \ref{appendix:rw}.

% We provide an analysis of previous methods in Appendix A and compare them with the proposed

\textbf{Input-side (Prompt-based) efficient reasoning.} These approaches enhance reasoning efficiency by controlling the input prompt, often based on task difficulty or explicit length constraints. For instance, Chain-of-Draft (CoD) \citep{xu2025chain} encourages step-by-step reasoning but restricts verbosity by requiring each step to be expressed in no more than five words. Similarly, Token-Budget \citep{han2024token} searches for optimal token budgets and incorporates them into prompts, thereby guiding the model to generate concise reasoning paths. 

\textbf{Model-side efficient reasoning.} This stream focuses on retraining or fine-tuning models to internalize more compact reasoning strategies. O1-Pruner \citep{luo2025o1} introduces a Length-Harmonizing Reward combined with a PPO-style optimization objective, enabling reasoning LLMs to produce shorter yet effective chains of thought (CoT). Similarly, TokenSkip \citep{xia2025tokenskip} constructs compressed CoT data by skipping less informative tokens and fine-tunes models on these shortened trajectories, thereby encouraging more efficient internal reasoning.

\textbf{Output-side efficient reasoning.} These methods dynamically shorten reasoning during inference by adjusting the generation process. Speculative Rejection \citep{sun2024fast} leverages a reward model to estimate partial sequence quality and terminates low-quality generations early. Early Stop Self-Consistency (ESC) \citep{li2024escape} instead monitors answer convergence within a sliding window, halting generation once outputs stabilize, thus preventing unnecessary reasoning steps.

\section{Method}

\subsection{Notation and Objective }
\label{sec:preliminary}

% ES-CoT is based on the hypothesis that an LLM tends to increase its confidence in a reasoning task as the generation progresses. 
For a target task with prompt $p_m$, let $P_M$ denote a pretrained LLM that receives the prompt and generates a solution step by step. We define a \textit{step} as a portion of the CoT that starts with a linguistic marker, such as "wait" and "alternatively", and ends with a new line character \citep{yang2025dynamic}. Let $T$ be the total number of steps, which is finite due to output length constraints in LLM generation. 

In ES-CoT, at each step $t \in \{1,2,...T\}$, we append the prompt \textit{"The final answer is"} to elicit the model's current answer (the orange column in Figure \ref{fig:framework}). We call this the \textit{step answer} and denote its distribution as $X_t$, with $X_T$ representing the distribution of the final answer. A sample of this distribution is written as $x_t\sim X_t$.
% Each $X_t$ may span multiple tokens. 
Let $\mathcal{A}$ be the answer space (the set of all possible values of $X_t$) with size $|\mathcal{A}|$.
Formally, $X_t$ should be understood as the distribution obtained by repeatedly sampling the model with the same prompt and recording the frequency of each distinct answer. This definition abstracts away from token-level likelihoods: although longer answers naturally receive lower token-level probabilities, they remain comparable to shorter answers under this frequency-based view. 
The objective of early-stopping CoT is to terminate at an intermediate step $t<T$ that yields an answer similar to the final step, i.e., $X_t \approx X_T$, while keeping $t$ as small as possible to reduce inference costs.

\subsection{CoT Answer Dynamics}
\label{sec:method:empirical}

% To examine the distribution dynamics of $X_t$ along with the reasoning process, we count the number of identical answers 

We now examine how the step answer distribution evolves as $t$ increases. Experiments are conducted with three LLMs on six mathematical and logical reasoning datasets. Unless otherwise noted, results are reported as averages across all datasets. Details of the datasets and models are provided in Section \ref{sec:setup}. 

% First, we establish that $X_t$ converges to $X_T$. Specifically, we plot the probability that the step answer matches the final answer, i.e., $P(X_t = X_T)$, against normalized step progress $t/T$ (Figure \ref{fig:empirical_prob}). As shown, $P(X_t = X_T)$ increases with $t$, indicating that LLMs progressively approach the final answer as reasoning unfolds.

\begin{figure}[tbp]
    \centering
    \begin{subfigure}[t]{0.48\linewidth}
        \centering
        \includegraphics[width=\linewidth]{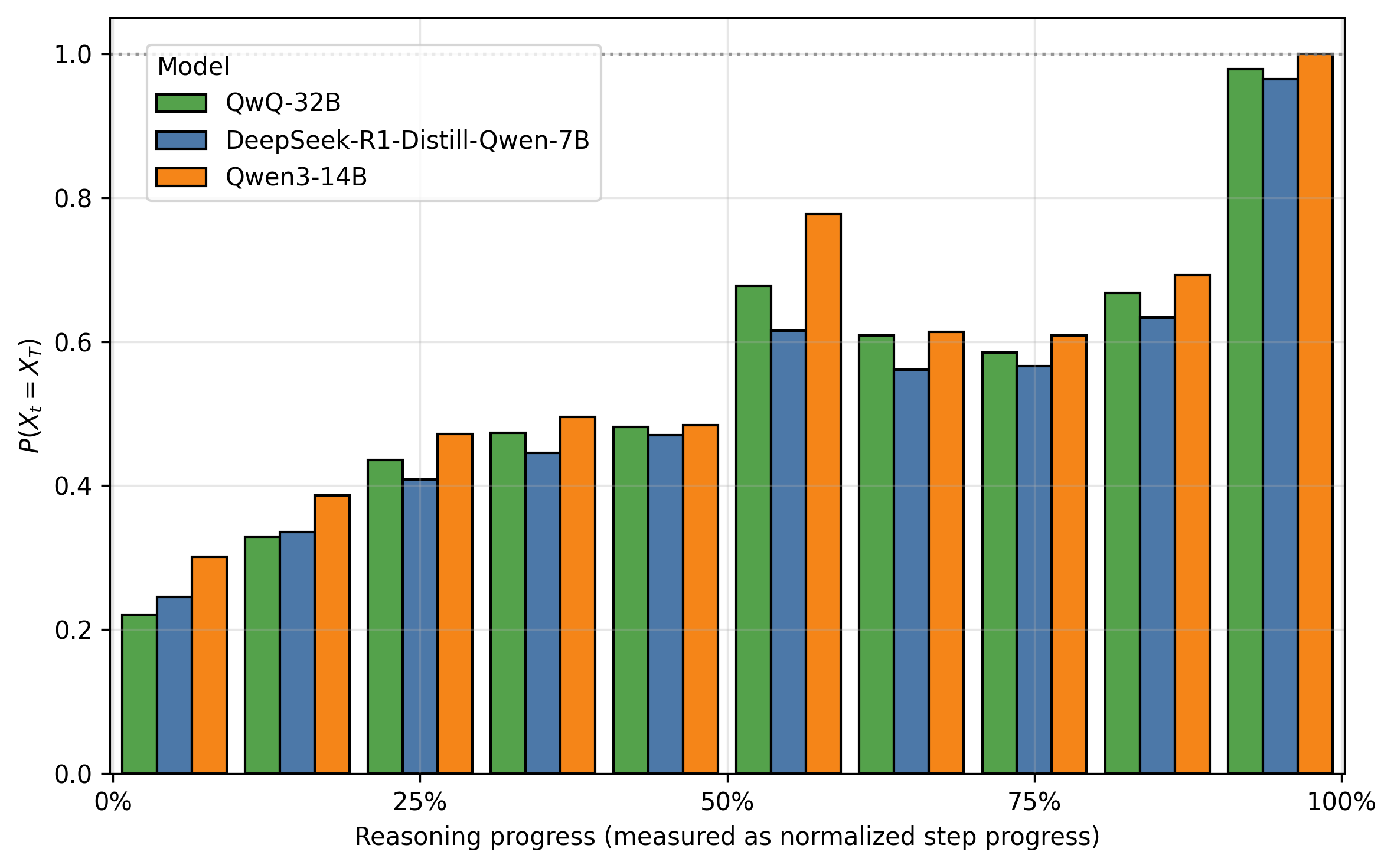}
        \caption{Probability that step answers match the final answer, $P(X_t=X_T)$, over reasoning progress ($t/T$). The last bar is an average over the final 10\% of steps, so its value can be less than 1.}
        \label{fig:empirical_prob}
    \end{subfigure}
    \hfill
    \begin{subfigure}[t]{0.48\linewidth}
        \centering
        \includegraphics[width=\linewidth]{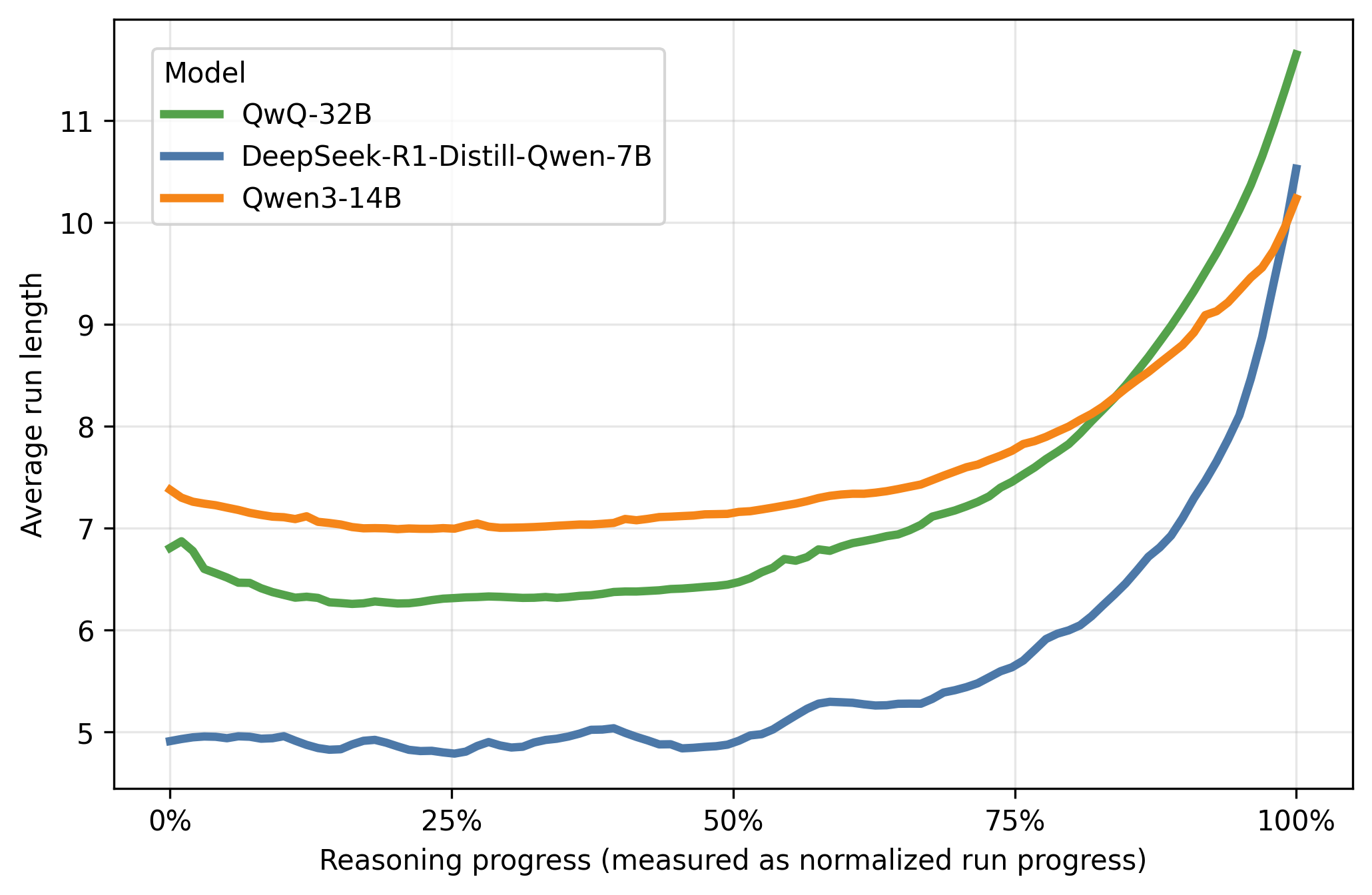}
        \caption{Run lengths over reasoning progress (measured as normalized run progress). Late-stage jumps indicate convergence.}
        \label{fig:empirical_count}
    \end{subfigure}
    \caption{CoT Answer Dynamics.}
    \label{fig:converge}
\end{figure}

We first conduct an empirical evaluation of the probability that the step answer matches the final answer, i.e., $P(X_t = X_T)$. We proceed as follows. (1) We first record the final answer $x_T$ for each CoT trajectory. (2) For each trajectory, we record the relative position $t/T$ whenever $x_t = x_T$. We then analyze the empirical density of these relative positions, which serves as a proxy for the probability $P(X_t = X_T)$. We depict the density distribution in Figure~\ref{fig:empirical_prob}. 
% As shown, $P(X_t = X_T)$ tends to increase with $t$, indicating that LLMs progressively approach the final answer throughout their reasoning process.
As shown, $P(X_t = X_T)$ increases with $t$, indicating that LLMs progressively approach the final answer as reasoning unfolds.

However, convergence alone does not provide a stopping criterion. Intuitively, $X_t$ stabilizes when multiple consecutive steps yield the same answer. To capture this, we measure the \textit{run length} of consecutive identical answers, and denote this sequence as $R=\langle r_1,r_2,\dots \rangle$. If $X_t$ is converging to $X_T$, we should observe an increasing $R$ as the model becomes more confident. For example, in Figure \ref{fig:framework}, early answers such as ``703'' or ``513'' appear only once, while ``197'' repeats eight times near the end. 

We plot the evolution of $R$ in Figure~\ref{fig:empirical_count}. Unlike Figure~\ref{fig:empirical_prob}, where the x-axis tracks step progress ($t/T$), here the x-axis tracks the index of each runs. For instance, in the run sequence $R=\langle 1,1,1,8 \rangle$, the first three short runs occupy 3/4 of the horizontal axis, while the long final run (``197'' repeated nine times) occupies the last 1/4. 
The averaged results across datasets and models show that run length grows as reasoning proceeds. More importantly, the growth curve is convex, suggesting a leap of convergence in later stages, where the model increasingly commits to a single answer. This late-stage convexity signals a decisive stopping criterion.

% \begin{figure}[htbp]
%     \centering
%     \includegraphics[width=0.8\linewidth]{fig/mean_runlength_by_model.png}
%     \caption{xxxx.}
%     \label{fig:empirical_count}
% \end{figure}

\begin{algorithm}[ht]
\caption{ES-CoT %(Run-jump Early Stop)
} 
\label{algo:es-cot}
\textbf{Input:} A predefined minimum difference $d_{min}$, a pretrained LLM, and a task represented by its prompt 
\begin{algorithmic}[1]
% \STATE Initialize an answer list $L_A$
\STATE Initialize $t=0$, a run sequence $R = \left< r_1 \right>$, and the run parameters $n=1$, $r_1=0$
\WHILE{not exceeding the reasoning budget}
\STATE Observe "wait", $t\gets t+1$
\STATE After the new line character, add the prompt \textit{`The final answer is'}, record the answer $x_t$
\IF{$x_t = x_{t-1}$ or $t=1$}
\STATE $r_n \gets r_n + 1$
\ELSE 
\STATE $n \gets n+1$ and $r_n = 1$
\ENDIF 
\STATE Update the run sequence $R$
\STATE Update the difference sequence $D = \left< d_1, d_2, \dots, d_{n-1} \right> = \left< r_k-r_{k-1} \right>_{k=2}^{n}$
\IF{$d_{n-1} \geq d_{min}$ and a t-test indicates that $d_{n-1}$ is significantly greater than previous differences $d_{1:n-2}$}
\STATE Terminate the generation, output the answer $x_t$
\ENDIF
\ENDWHILE
\STATE Output the final answer $x_T$
\end{algorithmic}
\textbf{Output: } The generated answer $x_t$ or $x_T$
\end{algorithm} 

\subsection{ES-CoT Algorithm %(Run-jump Criterion)
}

Building on the convergence patterns identified in Section~\ref{sec:method:empirical}, we formalize ES-CoT and examine its theoretical guarantees. Algorithm~\ref{algo:es-cot} specifies the procedure, while Section~\ref{sec:theory} develops conditions under which early stopping remains consistent with the final answer. 
%derived from Assumptions~\ref{assumption:1} and \ref{assumption:2}. 
Algorithm~\ref{algo:es-cot} maintains a run sequence $R=\left< r_1, r_2, \dots, r_{n}\right>$ that records consecutive identical answers. To monitor how quickly runs grow, we compute the difference sequence $D=\langle d_1, d_2, \dots, d_{n-1}\rangle$ with $d_i = r_{i+1} - r_i$. Early stopping is determined by the run-jump test: if the latest difference $d_{n-1}$ exceeds a predefined threshold $d_{min}$ and is statistically larger than the earlier differences according to a t-test, the reasoning is terminated. Otherwise, generation continues until completion. We formally define the run-jump test as follows.

\begin{definition}[Run-jump test]
Let $R=\langle r_1,\ldots,r_n \rangle $ be the run lengths and 
$D=\langle d_1,\ldots,d_{n-1} \rangle$ with $d_i = r_{i+1}-r_i$.
We trigger an early stop at run $n$ if
\[
  d_{n-1} \ge d_{\min}
  \quad\text{and}\quad
  d_{n-1}\ \text{is significantly larger than}\ d_{1{:}n-2}.
\]
\end{definition}

Intuitively, a sharp jump in run length signals a phase transition in the model’s confidence: the step answer has stabilized, and further reasoning adds little value. ES-CoT halts precisely at the convergence, achieving substantial savings in inference cost without sacrificing accuracy.

\subsection{Theoretical Analysis of ES-CoT}
\label{sec:theory}

We now analyze the theoretical properties of ES-CoT. Throughout, we adopt the definitions in Section~\ref{sec:preliminary}. Our focus is on regimes where the final step produces a confident answer. 
Tasks whose final predictions remain high-entropy are not amenable to early stopping in the first place.
This is because, for tasks with high-entropy answers, even the last step may not be a stopping point. Therefore, we make the following assumption.

% In this section, we provide a simple theoretical analysis of the efficacy of ES-CoT. Start from the definition in Section \ref{sec:preliminary}, without loss of generality, for an early stop CoT sequence, we assume the answer distribution $X_T$ at the last step is at low entropy, i.e., the LLM always gives a final answer at high confidence. This is because, for tasks with high-entropy answers, even the last step may not be a stopping point. Early stopping is not an issue for these tasks. Therefore, we make the following assumption.

\begin{assumption}
\label{assumption:1}
(\textbf{Deterministic final answer})
The final-step answer distribution is a Dirac delta. Writing $X_T=(p_T^1,\dots,p_T^{|\mathcal{A}|})$ over answer space $\mathcal{A}$, there exists an index $\mathrm{max}$ such that
\begin{equation}
    p_T^i = \begin{cases}
1 & \text{if}~i = max, \\
0 & \text{otherwise}.
\end{cases}
\end{equation}
% W.l.o.g., relabel so that $\mathrm{max}=1$.
W.l.o.g., let $\mathrm{max}=1$.
\end{assumption}

We empirically validate Assumption~\ref{assumption:1} in Appendix~\ref{appendix:ass1} by showing that LLMs consistently produce the same final answer under different random seeds.

Next, we make an assumption about how the step-answer distributions $X_t$ evolve during reasoning. Prior work shows that the mutual information between tokens near the end of each step and the final answer token increases as reasoning progresses \citep{qian2025demystifying}. Consistent with this, our empirical results in Figure~\ref{fig:empirical_prob} demonstrate that the probability $P(X_t = X_T)$ grows on average with $t$. Formally, we state the following assumption.

\begin{assumption}
\label{assumption:2}
(\textbf{Monotone approach to the final answer}) Let $p_t = P(X_t=X_T)$. We assume that $p_t$ is monotonically increasing in $t$. Under Assumption~\ref{assumption:1}, this is equivalent to $p_t^1$ increasing with $t$, since $X_T$ is a point mass on answer $1$ and $\Pr(X_t=X_T)=p_t^1$.

    % We assume that the probability of a step answer being the same as the final answer increases as generation proceeds. That is, let $p_t = P(X_t=X_T) $. We assume that $p_t$ increases monotonically with $t$. Note that under Assumption \ref{assumption:1}, it is equivalent that $p_t^1$ increases with $t$ because $X_T$ is a $\delta$ distribution and $P(X_t=X_T)=p_t^1$. 
\end{assumption}

Assumption \ref{assumption:1} and \ref{assumption:2} yield an explicit bound on the error of the answer obtained by ES-CoT mismatching the final answer. 

\begin{theorem}
\label{theorem:1}
    Let $k$ denote the index of the run where ES-CoT terminates early. Let $c_j=\sum_{i=1}^{j} r_i$ be the number of steps up to run $j$. For notation simplicity, denote $q=c_{k-1}$. Consider the $(k-1)$- and $k$-th runs. Let $e= P(X_{c_k} \neq X_T)$ be the error of ES-CoT, i.e., the intermediate answer obtained by ES-CoT does not match the final answer, then 
\begin{equation}
    e \leq 1 - \frac{1}{\left(\frac{1-p_{q+1}}{p_{q+1}}\right)^{(r_k-r_{k-1})} + 1}. 
\end{equation} 
Proof. See Appendix~\ref{appendix:proof}. 
\end{theorem}

\begin{remark}
\label{remark:1}
    Since $p_t$ is monotonically increasing and $p_T=1$, by the squeeze theorem, there exists a half step $h\leq T$, s.t., $p_t > \frac{1}{2}$ for any $t\geq h$. If $p_t > \frac{1}{2}$, then $\frac{1-p_{t}}{p_{t}} < 1$. With a high difference in the size of runs $r_k- r_{k-1}$, the error is approaching $0$. 
\end{remark}

Theorem~\ref{theorem:1} motivates the ES-CoT design: a \emph{large} positive jump $d_k=r_k-r_{k-1}$ is required before stopping. In Algorithm~\ref{algo:es-cot}, this is enforced by (i) a minimum threshold $d_{min}$ (a warmup that prevents premature stops when $p_t$ is still small), and (ii) a statistical test that the latest jump is unusually large relative to previous jumps.

% Therefore, in Algorithm \ref{algo:es-cot}, we set a minimum threshold of the difference $d_{min}$ and compare the current difference to previous differences. Only a significant increase in the difference can lead to an early stop. Meanwhile, the minimum threshold $d_{min}$ inherently ensures that the intermediate step cannot be arbitrarily small, serving as a `warmup' of the reasoning process. According to Remark~\ref{remark:1}, a reasonable warmup can guarantee that $p_t$ is not so small that an early stop occurs merely due to randomness. 

Theorem \ref{theorem:1} only considers the special case where the answer space contains only 2 answers. We next extend the analysis beyond binary answer spaces by adding a mild regularity assumption on the distribution of other answers. 

\begin{proposition}
\label{prop:main}
    Let the definition of $k, c_j, q$, and $e$ be the same as above. Suppose $|\mathcal{A}|\ge 3$ and, at each step, the distribution over \emph{incorrect} answers is uniform.\footnote{Formally, for each $t$, conditional on $X_t\neq A_1$, we assume $\Pr(X_t=A_i \mid X_t\neq A_1)=1/(|\mathcal{A}|-1)$ for all $i\neq 1$.}
    % We further assume the answer distribution, excluding $\mathcal{A}^1$, follows a uniform distribution. Then,
    The following inequality holds:
\begin{equation}
    e \leq 1 - \frac{1}{1 + (1-p_{q+1})^{(r_k-r_{k-1})} \left( \frac{2}{|\mathcal{A}|-1}\right) ^ {r_k} + \left(\frac{1-p_{q+1}}{p_{q+1}} \cdot \frac{1}{|\mathcal{A}|-1} \right)^{(r_k-r_{k-1})} }. 
\end{equation}
Proof. See Appendix~\ref{appendix:prop}.
\end{proposition}

\begin{remark}
    If $q+1$ exceeds the half step $h$ 
    %i.e., after the “majority step” 
    (where $p_{q+1}>\tfrac12$), the upper bound in Proposition~\ref{prop:main} goes to 0 as $r_k - r_{k-1} \to \infty$. 
\end{remark}

Taken together, Theorem~\ref{theorem:1} and Proposition~\ref{prop:main} formalize the intuition behind ES-CoT: once runs begin to lengthen quickly, the current answer is very likely to match the eventual final answer. Requiring a large, statistically significant jump in run length is therefore a principled stopping rule that trades a small, controllable error for substantial token savings.

\section{Experiments}

\subsection{Experimental Setup}
\label{sec:setup}

\textbf{Datasets.} We evaluate ES-CoT on six mathematical and formal logical reasoning datasets: the American Invitational Mathematics Examination (AIME24 and AIME25), GSM8K \citep{cobbe2021training}, AMC \citep{amc}, OlympiadBench (simplified as Olympiad for space concern) \citep{he2024olympiadbench}, and GPQA \citep{rein2024gpqa}. 
%We report the basic information and summary statistics of each dataset in Table \textcolor{red}{xxx}. 
For each dataset, we report two metrics: accuracy (\textbf{Acc.}) of generated answers and the average generated tokens (\textbf{Tok.}) \citep{kojima2022large}. Overall, we report the average accuracy across datasets and the average compression rate (CR) of each method relative to Naive across datasets. 

\textbf{Benchmark Methods.} We compare ES-CoT with other inference-stage efficient reasoning methods, including standard CoT prompting (Naive), chain of draft (CoD) \citep{xu2025chain}, compressed chain of thought (CCoT) \citep{cheng2024compressed}, NoThinking \citep{ma2025reasoning}, Dynasor \citep{fu2025reasoning}, and DEER \citep{yang2025dynamic}. 

Specifically, Naive evaluates the reasoning performance without any intervention. CoD prompts the LLM to use fewer tokens directly during prompting. CCoT performs reasoning with a sequence of learned embeddings. NoThinking ends the reasoning process at the first step by manually adding the end-of-thinking token. Dynasor prompts the LLM to get intermediate answers. We set Dynasor to prompt the LLM every 128 tokens and stop reasoning if ten consecutive identical answers are obtained. DEER monitors the confidence of the reasoning process and exits the reasoning process when the confidence exceeds a threshold.

\textbf{Implementation details.} We test ES-CoT and each benchmark method on three LLMs of different scales: DeepSeek-R1-Distill-Qwen-7B \citep{guo2025deepseek} (abbreviated as DeepSeek), Qwen-14B \citep{yang2025qwen3}, and QwQ-32B \citep{qwq32b}.
% ES-CoT is evaluated across three language models with varying scales: 
All models are prompted using the same zero-shot templates as \cite{kojima2022large}. Details on the prompts are provided in Appendix~\ref{appendix:prompt}.
% Hyperparameters are tuned to balance diversity and stability: 
We fix the temperature hyperparameter at 0.6. For QwQ-32B, we additionally apply top-$p=0.9$ and top-$k=20$, while smaller models (DeepSeek-R1-Distill-Qwen-7B and Qwen-14B) decode without truncating. For ES-CoT, we set the minimum run-length difference to $d_{min}=20$, and determine significance using a $t$-test with a significance p-value of 0.1. 

% \textbf{Evaluation protocol.} 

We assess ES-CoT in three stages. First, we compare it directly to baselines (Section~\ref{sec:result:main}). Second, we perform sensitivity analyses (Section~\ref{sec:sensitivity}) to study the robustness of ES-CoT with respect to hyperparameters. Third, we discuss the results of ES-CoT in detail by investigating how ES-CoT significantly saves tokens.

\begin{table}[ht]
    % \centering
    % \small
    \setlength{\tabcolsep}{2.3pt}
    \begin{tabular}{l|cc|cc|cc|cc|cc|cc|cc}
    \toprule
     & \multicolumn{10}{c}{Math} & \multicolumn{2}{c}{Science}\\ 
     & \multicolumn{2}{c}{AIME24} & \multicolumn{2}{c}{AIME25} & \multicolumn{2}{c}{GSM8K} & \multicolumn{2}{c}{AMC} & \multicolumn{2}{c}{Olympiad} & \multicolumn{2}{c}{GPQA} & \multicolumn{2}{c}{Overall}\\ 
    Method & Acc & Tok & Acc & Tok & Acc & Tok & Acc & Tok & Acc & Tok & Acc & Tok & Acc & CR(\%) \\ 
    \midrule
    \multicolumn{15}{l}{\textit{\textbf{LLM: DeepSeek-R1-Distill-Qwen-7B}}} \\ 
    \rowcolor{gray!20} Naive & 0.60 & 11.29 & 0.37 & 11.81 & 0.91 & 1.45 & 0.88 & 6.45 & 0.57 & 8.44 & 0.45 & 9.86 & 0.63 & 100 \\ 
    CoD & 0.47 & 10.80 & 0.33 & 10.42 & 0.85 & 0.31 & 0.90 & 4.07 & 0.50 & 6.23 & 0.53 & 6.33 & 0.60 & 67.73 \\ 
    CCoT & 0.50 & 10.53 & 0.37 & 11.18 & 0.87 & 0.48 & 0.88 & 5.71 & 0.52 & 7.80 & 0.47 & 9.05 & 0.60 & 82.29 \\ 
    NoThinking & 0.13 & 4.24 & 0.23 & 3.92 & 0.87 & 0.23 & 0.60 & 2.96 & 0.41 & 1.97 & 0.39 & 1.06 & 0.44 & \underline{\textbf{27.77}} \\ 
    % TCC & 0.43 & 6.24 & 0.40 & 6.40 & 0.87 & 0.29 & 0.90 & 2.05 & 0.85 & 3.50 & 0.55 & 4.65 & 0.46 & 5.57\\ 
    Dynasor & 0.27 & 3.25 & 0.20 & 2.98 & 0.88 & 0.70 & 0.60 & 2.36 & 0.32 & 1.98 & 0.39 & 1.38 & 0.44 & 29.39 \\
    DEER & 0.53 & 8.89 & 0.37 & 11.37 & 0.91 & 0.70 & 0.85 & 4.62 & 0.51 & 6.74 & 0.26 & 9.79 & 0.57 & 79.01 \\ 
    \rowcolor{lightblue!20} ES-CoT & 0.60 & 9.51 & 0.33 & 8.75 & 0.91 & 1.22 & 0.83 & 5.15 & 0.48 & 6.55 & 0.53 & 6.31 & \underline{\textbf{0.61}} & 77.32 \\
    \midrule
    \multicolumn{15}{l}{\textit{\textbf{LLM: Qwen-14B}}} \\ 
    \rowcolor{gray!20} Naive & 0.73 & 11.35 & 0.73 & 12.80 & 0.96 & 1.65 & 0.98 & 7.34 & 0.67 & 8.81 & 0.65 & 7.32 & 0.79 & 100 \\ 
    CoD & 0.70 & 9.73 & 0.63 & 11.60 & 0.96 & 0.67 & 0.95 & 5.35 & 0.68 & 7.03 & 0.60 & 4.43 & 0.75 & 71.69 \\ 
    CCoT & 0.70 & 11.34 & 0.63 & 11.90 & 0.96 & 1.02 & 0.95 & 6.59 & 0.67 & 7.94 & 0.62 & 6.10 & 0.76 & 86.32 \\ 
    NoThinking & 0.23 & 7.82 & 0.20 & 5.87 & 0.95 & 0.25 & 0.70 & 2.86 & 0.52 & 3.42 & 0.56 & 2.47 & 0.53 & \underline{\textbf{40.24}} \\ 
    % TCC & 0.73 & 6.76 &0.50 & 7.52 & 0.96 & 1.01 &0.95 & 2.77 & 0.93 & 4.33 &0.68 & 5.21 & 0.60 & 4.13 \\ 
    Dynasor & 0.27 & 7.00 & 0.27 & 6.63 & 0.95 & 1.20 & 0.68 & 3.49 & 0.48 & 4.54 & 0.43 & 1.45 & 0.51 & 50.85 \\
    DEER & 0.63 & 8.64 & 0.57 & 10.91 & 0.95 & 0.69 & 0.93 & 4.27 & 0.59 & 5.00 & 0.63 & 7.06 & 0.72 & 69.09 \\

    \rowcolor{lightblue!20} ES-CoT & 0.73 & 10.78 & 0.67 & 10.56 & 0.96 & 1.62 & 0.95 & 6.55 & 0.65 & 7.86 & 0.64 & 6.77 & \underline{\textbf{0.77}} & 91.10 \\
    \midrule
    \multicolumn{15}{l}{\textit{\textbf{LLM: QwQ-32B}}} \\ 
    \rowcolor{gray!20} Naive & 0.67 & 11.14 & 0.63 & 12.03 & 0.96 & 1.43 & 0.93 & 6.80 & 0.69 & 8.16 & 0.69 & 7.45 & 0.76 & 100 \\ 
    CoD & 0.67 & 9.66 & 0.53 & 11.34 & 0.96 & 0.60 & 0.93 & 5.90 & 0.70 & 6.89 & 0.63 & 5.72 & 0.74 & 78.49 \\ 
    CCoT & 0.67 & 10.01 & 0.53 & 11.39 & 0.97 & 0.94 & 0.93 & 6.64 & 0.70 & 7.32 & 0.67 & 6.14 & \underline{\textbf{0.75}} & 86.67 \\ 
    NoThinking & 0.67 & 10.75 & 0.63 & 11.76 & 0.96 & 1.55 & 0.88 & 7.05 & 0.69 & 8.18 & 0.66 & 8.09 & \underline{\textbf{0.75}} & 100.25 \\ 
    % TCC & 0.77 & 6.15 & 0.60 & 6.35 & 0.96 & 0.98 & 0.95 & 2.27 & 0.93 & 4.00 & 0.68 & 4.95 & 0.67 & 4.51\\ 
    Dynasor & 0.37 & 9.30 & 0.17 & 11.51 & 0.96 & 1.18 & 0.70 & 6.52 & 0.49 & 7.30 & 0.40 & 4.71 & 0.52 & 85.04 \\
    DEER & 0.67 & 10.94 & 0.63 & 10.95 & 0.96 & 0.96 & 0.90 & 6.37 & 0.66 & 4.02 & 0.63 & 4.25 & 0.74 & \underline{\textbf{76.06}} \\ 
    \rowcolor{lightblue!20} ES-CoT & 0.67 & 8.77 & 0.60 & 9.77 & 0.96 & 1.39 & 0.93 & 5.98 & 0.66 & 6.48 & 0.66 & 5.63 & \underline{\textbf{0.75}} & 83.34 \\
    \midrule
    \end{tabular}
    \caption{Accuracy and average number of generated tokens (in thousands) across 6 datasets and 3 LLMs. 
    % We denote the average compression rate of each method relative to Naive across datasets as CR. 
    We bold and underline the results with the best overall accuracy and CR, excluding Naive. Note that the number of tokens contributed by the manually added prompts in ES-CoT is also counted.}
    \label{tab:result:main}
\end{table}

\subsection{Main Results}
\label{sec:result:main}

\textbf{Token efficiency and accuracy.} Table \ref{tab:result:main} reports accuracy and the average number of generated tokens for our method and each benchmark method.   
We begin by comparing ES-CoT against standard CoT prompting (Naive) with greedy decoding \citep{wei2022chain}. ES-CoT significantly reduces costs while barely affecting performance. As shown, compared to Naive, ES-CoT reduces token usage by an average of 8.9\% to 22.68\%. 
For example, on the GPQA dataset with DeepSeek, Naive requires 9,860 tokens per answer on average, whereas ES-CoT only needs 6,310 tokens, resulting in a reduction of 36\%. These results demonstrate that ES-CoT delivers substantial savings in inference cost by shortening reasoning traces. 
Turning to accuracy, ES-CoT achieves performance comparable to Naive across all tasks. In some cases, accuracy even improves: on the GPQA dataset with DeepSeek, ES-CoT surpasses CoT despite using fewer tokens. This suggests that ES-CoT not only reduces cost but occasionally mitigates overthinking, improving answer quality \citep{chen2024not}. Overall, the results in Table~\ref{tab:result:main} show that ES-CoT maintains accuracy while significantly lowering inference cost.

Next, we compare our method to recent inference-stage efficient reasoning methods. Among them, CoD, CCoT, and NoThinking are prompt-based methods that directly control the reasoning trace by modifying prompts. Dynasor and DEER monitor reasoning dynamics and exit early when meeting certain conditions. It can be observed that ES-CoT consistently outperforms all baselines in terms of accuracy. While NoThinking and CCoT achieve similar results with QwQ, their performance does not generalize well to other LLMs. For example, NoThinking achieves an average accuracy of 0.44 with DeepSeek and 0.53 with Qwen. 
Moreover, for prompt-based methods, the results are not easily controllable. However, we demonstrate the flexibility of our approach in Section \ref{sec:sensitivity}. By adjusting the parameters of our method, we can achieve different trade-offs between accuracy and token usage. For baselines that monitor reasoning dynamics (including Dynasor and DEER), note that accuracy is often greatly sacrificed. For example, Dynasor achieves only an overall accuracy of 0.44 with DeepSeek, and DEER achieves 0.57. This indicates that these baselines struggle to identify the correct stopping criteria for reasoning. Notably, ES-CoT is a general method that does not require any prior knowledge of model capabilities or task difficulty. The cost reduction with minimal impact on correctness stems directly from the early-stop mechanism.

\subsection{Sensitivity Analysis }
\label{sec:sensitivity}

We next study the sensitivity of ES-CoT to its two hyperparameters: the minimum run-length difference $d_{min}$ and the significance level of the t-test.

\begin{table}[ht]
    \centering
    \setlength{\tabcolsep}{6pt}
    \begin{tabular}{lcccccc}
        \toprule
         $d_{min}$ & 5 & 10 & 15 & 20 & 30 & 40  \\
         \midrule
         \multicolumn{7}{l}{\textit{\textbf{DeepSeek-R1-Distill-Qwen-7B}}} \\
         Acc$\uparrow$ & 0.56 & 0.60 & 0.61 & 0.61 & 0.62 & 0.62 \\
         CR $\downarrow$ & 59.00 & 68.23 & 73.42 & 77.32 & 82.12 & 85.47 \\
         \midrule
         \multicolumn{7}{l}{\textit{\textbf{Qwen-14B}}} \\
         Acc$\uparrow$ & 0.67 & 0.72 & 0.75 & 0.77 & 0.77 & 0.77 \\
         CR $\downarrow$ & 69.11 & 79.19 & 86.55 & 91.10 & 95.70 & 97.64 \\
         \midrule
         \multicolumn{7}{l}{\textit{\textbf{QwQ-32B}}} \\
         Acc$\uparrow$ & 0.67 & 0.71 & 0.74 & 0.75 & 0.76 & 0.76 \\
         CR $\downarrow$ & 63.75 & 73.32 & 79.33 & 83.34 & 89.93 & 93.25 \\
         \bottomrule
    \end{tabular}
    \caption{Average accuracy and overall compression rate with different $d_{min}$ across 6 datasets. Detailed results are reported in Table \ref{tab:dmin}.}
    \label{tab:sensitivity}
\end{table}

\textbf{Effect of $d_{min}$.}
Table \ref{tab:sensitivity} reports accuracy and overall compression rate for different values of $d_{min}$ with a fixed p-value of 0.1. We also report detailed results for each dataset with different values of $d_{min}$ in Table \ref{tab:dmin} in Appendix \ref{appendix:dmin}. It is shown both in Table \ref{tab:sensitivity} and \ref{tab:dmin} that both accuracy and compression rate rise steadily as $d_{min}$ grows. Intuitively, a larger threshold delays early stopping, allowing more steps before termination. 
In the limit ($d_{min}=40$), ES-CoT can match the accuracy of full CoT prompting, albeit at the cost of generating more tokens. For example, with QwQ, a value of $d_{min}=30$ achieves nearly the same overall accuracy as Naive while still reducing token usage by 6.75\%. On the other hand, when $d_{min}=5$, ES-CoT achieves a large token reduction, with a maximum overall CR of 69.11\%, indicating that 30.89\% are reduced. This demonstrates that ES-CoT behaves as a scalable decoding procedure: small $d_{min}$ favors efficiency, while larger $d_{min}$ favors accuracy. In comparison to prompt-based baselines (e.g., CoD, CCoT, and NoThinking), ES-CoT offers the flexibility to balance accuracy and token usage by simply adjusting the parameter.

% \begin{figure}[ht]
%     \centering
%     \includegraphics[width=\linewidth]{fig/pvalue_figs/amc_average_tokens.png}
%     \caption{Robustness analysis of ES-CoT regarding the hyperparameters, including the minimum difference $d_{min}$ and p-value. For each line, we fix the p-value and vary $d_{min}$ to get the results. The results are calculated on AIME25 with different LLMs.}
%     \label{fig:p-value}
% \end{figure}

\textbf{Effect of the t-test significance level.} We also vary the p-values threshold and analyze its impact. Specifically, we vary the p-value from 0.05 to 0.2. For each fixed p-value, we set $d_{min}$ to range from 5 to 40. A lower p-value enforces stricter statistical evidence, requiring larger jumps in run length to trigger early stopping. For space concerns, we report the results in Figure \ref{fig:pvalue1} and \ref{fig:pvalue2} in Appendix \ref{appendix:additional_results}. 
% Figure \ref{fig:p-value} shows the tradeoff between accuracy and the number of tokens on AIME25 across three LLMs. 
As shown, for each LLM, the lines in different colors do not exhibit a significant difference. This demonstrates that ES-CoT remains robust across a wide range of p-values, indicating $d_{min}$ is the dominant parameter governing the cost-accuracy balance. 
% We refer readers to Appendix~\ref{appendix:robust} for results on additional datasets, where we reach similar conclusions across all datasets and LLMs.

% Combining Table \ref{tab:result:main} and Table \ref{tab:sensitivity}, we also observe that the number of tokens is positively correlated with performance, which means a high inference cost (more tokens) can lead to a high accuracy. 

% \textbf{Cost saving are positively correlated with performance.} It is intuitive since better performance often eliminates the need for longer thought tokens. 

\subsection{Discussion }

\begin{table}[ht]
    \centering
    \begin{tabular}{lcccccc}
        \toprule
         & AIME24 & AIME25 & GSM8K  & AMC  & Olympiad & GPQA\\
         \midrule
         DeepSeek & 0.87 & 0.80 & 0.99 & 0.93 & 0.78 & 0.81\\
         Qwen & 0.93 & 0.87 & 1.00 & 0.95 & 0.92 & 0.97 \\
         QwQ & 0.97 & 0.87 & 1.00 & 1.00 & 0.88 & 0.93\\
         \bottomrule
    \end{tabular}
    \caption{The ratio of instances where ES-CoT and Naive produce the same answer.}
    \label{tab:same}
\end{table}

\textbf{Intersection of ES-CoT and Naive.} Since the objective of ES-CoT is to stop reasoning early while preserving the final answer, we further examine the overlap between ES-CoT and Naive outputs. Table~\ref{tab:same} reports the ratio of instances where the two methods produce the same answer. Consistent with our theoretical analysis in Section~\ref{sec:theory}, the overlap ratios are high across all models and datasets. In particular, Qwen and QwQ yield at least 87\% identical answers, confirming that ES-CoT typically halts at the point where the final answer has already stabilized.

\textbf{Case study.} We provide a case study in Figure \ref{fig:case} in Appendix. This example shows that, for a typical question, the LLM reaches a confident and correct result at an early stage of reasoning. However, without early stopping, the model would continue to generate 3,701 tokens additionally to confirm the final answer. In contrast, based on the early stop mechanism of ES-CoT, the model stops the generation after using only 1,579 tokens. Our method is simple and effective in avoiding the overthinking issue of the model. 

\section{Conclusions }

In this study, we introduce ES-CoT, an inference-time method that shortens chain-of-thought reasoning while preserving answer quality. ES-CoT tracks runs of identical step answers and halts generation when the most recent run exhibits a statistically significant leap beyond prior runs and exceeds a minimum threshold. We provide empirical evidence that two patterns consistently emerge in reasoning models: (i) run length grows as reasoning proceeds, and (ii) the probability that a step answer matches the final answer increases along the trajectory. Building on these observations, we present a theoretical analysis that explains why a large jump in run length is a reliable signal for termination. 
Experiments on five reasoning benchmarks and three models confirmed that ES-CoT reduces token usage by about 16.08\% on average while maintaining similar accuracy. 

% We also demonstrate that ES-CoT integrates well with self-consistency and remains robust across a wide range of hyperparameters.

\section{Future Work}

Looking ahead, several extensions are promising. One extension is to make ES-CoT more adaptive. For example, future work can first adjust the minimum difference and significance level based on instance-level features, then employ ES-CoT for efficient reasoning. Another direction is to broaden the evaluation to closed-source models and domains beyond mathematics and formal logic. 
% A further extension is to incorporate measures of uncertainty. As the present approach only counts step answers without assessing the quality or confidence of each step, integrating uncertainty estimates could yield more reliable stopping decisions. 
% An additional extension is to consider tasks without deterministic answers. Reducing token usage in uncertain problems remains a challenge. 
Finally, while ES-CoT currently focuses on convergence to the model’s own final prediction, an important direction is to investigate whether early stopping can be guided by signals more directly related to the ground truth, especially in cases where the model’s final prediction is incorrect.

\bibliography{colm2026_conference}
\bibliographystyle{colm2026_conference}

\appendix
% \section{Appendix}

\section{Research Gap Summary}
\label{appendix:rw}

Input-side prompting methods rely on problem-specific analysis to achieve better performance, while model-side retraining requires additional training or fine-tuning of LLMs, which is often costly. In contrast, our approach belongs to output-side efficient reasoning: it adjusts reasoning length across tasks and models without extra supervision. Compared to other output-side methods \citep{sun2024fast,li2024escape}, ES-CoT avoids parallel decoding and eliminates the needs for auxiliary reward models. Building on empirical evidence of LLM reasoning dynamics, ES-CoT introduces the run-jump test, a simple and single-trajectory rule on answer run lengths that early stops the reasoning when the current run makes a statistically significant jump. 

\section{Empirical Validation of Assumption 1}
\label{appendix:ass1}

We empirically validate Assumption~\ref{assumption:1} that the final answer distribution \(X_T\) is sharply concentrated. 
For each problem instance, we fix the prompt and the CoT setup, then sample the model’s final answer \(X_T\) ten times with temperature \(0.6\) and top-p 0.9. 
We take the greedy-decoded final answer as the reference and compute its share among the ten samples. 
We then average this share over all instances within each dataset and report the averages for each model–dataset pair.

If Assumption 1 holds, the reference share should be high and close to one, indicating that the model’s final answer is stable under modest stochastic sampling. 
As shown in Table~\ref{tab:assumption1}, the averages are consistently high across three models and five datasets. 
This pattern indicates strong answer-level concentration of \(X_T\) and provides direct empirical support for Assumption~\ref{assumption:1}.

% \begin{table}[htbp]
%     \centering
%     \begin{tabular}{cccccc}
%         \toprule
%          & AIME & GPQA & MATH & Minerva & Olympiad \\
%          \midrule
%          QwQ & 1.00 & 0.98 & 0.86 & 0.85 & 0.82 \\
%          Qwen3 & 1.00 & 0.99 & 0.90 & 0.90 & 0.89 \\
%          DeepSeek & 1.00 & 0.98 & 0.87 & 0.81 & 0.83 \\
%          \bottomrule
%     \end{tabular}
%     \caption{Average proportion of the greedy-decoded final answer among 10 sampled final answers (temperature $=0.6$, top-$p=0.9$) across datasets and models. Higher values indicate stronger stability of $X_T$, providing empirical support for Assumption~\ref{assumption:1}.}
%     \label{tab:assumption1}
% \end{table}

\begin{table}[ht]
    \centering
    \begin{tabular}{lcccccc}
    \toprule
     & AIME24 & AIME25 & GSM8K  & AMC  & Olympiad & GPQA\\
     \midrule
     DeepSeek & 0.70 & 0.71 & 0.93 & 0.95 & 0.75 & 0.77\\
     Qwen & 0.88 & 0.86 & 1.00 & 0.98 & 0.91 & 0.96\\
     QwQ & 0.92 & 0.80 & 1.00 & 0.95 & 0.87 & 0.94\\
     \bottomrule
    \end{tabular}
    \caption{Average proportion of the greedy-decoded final answer among 10 sampled final answers (temperature $=0.6$, top-$p=0.9$) across datasets and models. Higher values indicate stronger stability of $X_T$, providing empirical support for Assumption~\ref{assumption:1}.}
    \label{tab:assumption1}
\end{table}

\section{Proofs}

\subsection{Proofs of Theorem 1}
\label{appendix:proof}

\begin{theorem_app} 
    Let $k$ denote the index of the run where ES-CoT terminates early. Let $c_j=\sum_{i=1}^{j} r_i$ be the number of steps in the first $j$ runs. For notation simplicity, denote $q=c_{k-1}$. Consider answers in the $(k-1)$- and $k$-th runs. Let $e$ be the error of ES-CoT, i.e., the intermediate answer obtained by ES-CoT does not match the final answer $e = P(X_{c_k} \neq X_T)$. If $|\mathcal{A}|=2$, then 
\begin{equation}
    e \leq 1 - \frac{1}{\left(\frac{1-p_{q+1}}{p_{q+1}}\right)^{(r_k-r_{k-1})} + 1}. 
\end{equation} 
\end{theorem_app}

\textbf{Proof.} Let $\mathcal{A}^i$ denote the $i$-th answer in the answer space, where in this case $i\in \{1, 2\}$. By definition, in the $(k-1)$-th run, one answer occurs $r_{k-1}$ times, while in the $k$-th run, a different answer occurs $r_k$ times. According to Assumption \ref{assumption:1}, $X_T = (1,0)$ and we want $\mathcal{A}^1$ to be generated. Consider the posterior distribution after the observation, where one answer occurs $r_{k-1}$ times, then another answer occurs $r_k$ times. In the case where $|\mathcal{A}|=2$, the only possible outcomes are $\left(\mathcal{A}^1,\mathcal{A}^2\right)$ and $\left(\mathcal{A}^2,\mathcal{A}^1\right)$. By Assumption \ref{assumption:2}, we have 
\begin{equation}
\label{eq:proof1}
    P\left(\left(\mathcal{A}^1,\mathcal{A}^2\right)\right) \leq p_{c_{k-1}}^{r_{k-1}} \cdot (1-p_{c_{k-1}+1})^{r_k}, 
\end{equation}
where $p_t$ is defined in Assumption \ref{assumption:2}. This is because $p_t$ increases with $t$. For generating $\mathcal{A}^1$, the maximum probability is at the last step $c_{k-1}$. For generating $\mathcal{A}^2$, the maximum probability is at the first step $c_{k-1}+1$. For simplicity, let $q=c_{k-1}$. Inequality \ref{eq:proof1} can be re-written as 
\begin{equation}
\label{eq:ge1}
    P\left(\left(\mathcal{A}^1,\mathcal{A}^2\right)\right) \leq p_{q}^{r_{k-1}} \cdot (1-p_{q+1})^{r_k}.
\end{equation}
% which is because $p_q \leq p_{q+1}$.

Similarly, for the case of $\left(\mathcal{A}^2,\mathcal{A}^1\right)$, consider the minimum probability and we have 
\begin{equation}
\label{eq:ge2}
    P\left(\left(\mathcal{A}^2,\mathcal{A}^1\right)\right) \geq (1-p_{q})^{r_{k-1}} \cdot (p_{q+1})^{r_k}.
\end{equation}

With the observation, we have the conditional probability of $\left(\mathcal{A}^2,\mathcal{A}^1\right)$, 
\begin{equation}
\begin{aligned}
P\left(\left(\mathcal{A}^2,\mathcal{A}^1\right)|\text{Obs}\right) & = \frac{P\left(\left(\mathcal{A}^2,\mathcal{A}^1\right)\right)}{P\left(\left(\mathcal{A}^1,\mathcal{A}^2\right)\right) + P\left(\left(\mathcal{A}^2,\mathcal{A}^1\right)\right)} \\ 
& \geq \frac{(1-p_{q})^{r_{k-1}} \cdot (p_{q+1})^{r_k}}{p_{q}^{r_{k-1}} \cdot (1-p_{q+1})^{r_k} + (1-p_{q})^{r_{k-1}} \cdot (p_{q+1})^{r_k}} \\
& = \frac{1}{(\frac{p_q}{1-p_q})^{r_{k-1}} (\frac{1-p_{q+1}}{p_{q+1}})^{r_k} + 1} \\ 
& \geq \frac{1}{\left(\frac{1-p_{q+1}}{p_{q+1}}\right)^{(r_k-r_{k-1})} + 1}.
\end{aligned}
\end{equation}
The second step is because of Inequalities \ref{eq:ge1} and \ref{eq:ge2}. The fourth step is because $\frac{p_q}{1-p_q} \leq \frac{p_{q+1}}{1-p_{q+1}}$. 

Finally, the error is defined as 
\begin{equation}
    e = 1- P\left(\left(\mathcal{A}^2,\mathcal{A}^1\right)|\text{Obs}\right).
\end{equation}

This completes the proof. 

\subsection{Proofs of Proposition 1}
\label{appendix:prop}

\begin{proposition_app}
\label{prop:1}
    Let $k$ denote the index of the run where ES-CoT terminates early. Let $c_j=\sum_{i=1}^{j} r_i$ be the number of steps in the first $j$ runs and $q=c_{k-1}$. Let $e$ be the error of ES-CoT, i.e., the intermediate answer obtained by ES-CoT does not match the final answer $e = P(X_{c_k} \neq X_T)$. We further assume the answer distribution, excluding $\mathcal{A}^1$, follows a uniform distribution. Then, the following holds:
\begin{equation}
    e \leq 1 - \frac{1}{1 + (1-p_{q+1})^{r_k-r_{k-1}} \left( \frac{2}{|\mathcal{A}|-1}\right) ^ {r_k} + \left(\frac{1-p_{q+1}}{p_{q+1}} \cdot \frac{1}{|\mathcal{A}|-1} \right)^{r_k-r_{k-1}} }. 
\end{equation}
\end{proposition_app}

\textbf{Proof.} Similarly, we consider the posterior distribution for the observation in the $(k-1)$- and $k$-th runs. The possible outcomes are now $\left(\mathcal{A}^i,\mathcal{A}^1\right)$, $\left(\mathcal{A}^i,\mathcal{A}^j\right)$, and $\left(\mathcal{A}^1,\mathcal{A}^i\right)$, where $i\neq j$, $i\neq 1$, and $j\neq 1$. With the additional uniform assumption, we have 
\begin{equation}
\label{eq:ge3}
    P\left(\left(\mathcal{A}^i,\mathcal{A}^1\right)\right) \geq \left(\frac{1-p_{q}}{|\mathcal{A}|-1}\right)^{r_{k-1}} \cdot (p_{q+1})^{r_k},
\end{equation}
\begin{equation}
\label{eq:ge4}
    P\left(\left(\mathcal{A}^i,\mathcal{A}^j\right)\right) \leq \left(\frac{1-p_{q-r_{k-1}}}{|\mathcal{A}|-1}\right)^{r_{k-1}} \cdot \left(\frac{1-p_{q+1}}{|\mathcal{A}|-1}\right)^{r_k},
\end{equation}
\begin{equation}
\label{eq:ge5}
    P\left(\left(\mathcal{A}^1,\mathcal{A}^i\right)\right) \leq p_q^{r_{k-1}} \cdot \left(\frac{1-p_{q+1}}{|\mathcal{A}|-1}\right)^{r_k}.
\end{equation}

With the observation, we have the conditional probability of $\left(\mathcal{A}^i,\mathcal{A}^1\right)$, 
\begin{equation}
\begin{aligned}
P & \left(\left(\mathcal{A}^i,\mathcal{A}^1\right)|\text{Obs}\right)  = \frac{P\left(\left(\mathcal{A}^i,\mathcal{A}^1\right)\right)}{P\left(\left(\mathcal{A}^i,\mathcal{A}^1\right)\right) + P\left(\left(\mathcal{A}^i,\mathcal{A}^j\right)\right) + P\left(\left(\mathcal{A}^1,\mathcal{A}^i\right)\right)} \\ 
& \geq \frac{1}{1+ \left( \frac{1-p_{q-r_{k-1}}}{1-p_{q}} \right) ^ {r_{k-1}}  \left(\frac{1-p_{q+1}}{p_{q+1}} \right)^{r_k}  \left( \frac{1}{|\mathcal{A}|-1}\right) ^ {r_k} +  \left(\frac{p_q}{1-p_q}\right)^{r_{k-1}} \left(\frac{1-p_{q+1}}{p_{q+1}} \right)^{r_k}  \left( \frac{1}{|\mathcal{A}|-1}\right) ^ {r_k-r_{k-1}} }.
\end{aligned}
\end{equation}
Let $q$ be large enough to exceed the half step $h$ (see Remark \ref{remark:1} for the definition of $h$). Then, for the first term in the denominator, we have 
\begin{equation}
\begin{aligned}
    \left( \frac{1-p_{q-r_{k-1}}}{1-p_{q}} \right) ^ {r_{k-1}} & \left(\frac{1-p_{q+1}}{p_{q+1}} \right)^{r_k}  \left( \frac{1}{|\mathcal{A}|-1}\right) ^ {r_k} \\
    & \leq \left( \frac{1}{1-p_{q}} \right) ^ {r_{k-1}} \left(\frac{1-p_{q+1}}{p_{q+1}} \right)^{r_k}  \left( \frac{1}{|\mathcal{A}|-1}\right) ^ {r_k} \\
    & = \left( \frac{1-p_{q+1}}{1-p_{q}} \right) ^ {r_{k-1}} (1-p_{q+1})^{r_k-r_{k-1}}  \left(\frac{1}{p_{q+1}} \right)^{r_k}  \left( \frac{1}{|\mathcal{A}|-1}\right) ^ {r_k} \\
    & \leq 1 \cdot (1-p_{q+1})^{r_k-r_{k-1}} \left( \frac{2}{|\mathcal{A}|-1}\right) ^ {r_k}.
\end{aligned}    
\end{equation}
The first step is because $p_{q-r_{k-1}}\geq 0$. The third step is because $1-p_{q+1}\leq 1-p_q$ and $p_{q+1}\geq 0.5$. Note that, when $|\mathcal{A}| > 3$, this term is approaching 0 when $p_{q+1}$ is approaching 1 or ${r_k-r_{k-1}}$ is approaching $\infty$. 

Next, for another term in the denominator, similar to the proof in Theorem \ref{theorem:1}, we have 
\begin{equation}
\begin{aligned}
    \left(\frac{p_q}{1-p_q}\right)^{r_{k-1}} & \left(\frac{1-p_{q+1}}{p_{q+1}} \right)^{r_k}  \left( \frac{1}{|\mathcal{A}|-1}\right) ^ {r_k-r_{k-1}} \leq \left(\frac{1-p_{q+1}}{p_{q+1}}\right)^{r_k-r_{k-1}} \left( \frac{1}{|\mathcal{A}|-1}\right) ^ {r_k-r_{k-1}}.
\end{aligned}    
\end{equation}

When $|\mathcal{A}| > 2$, if $p_{q+1}>0.5$, the term is approaching 0 when ${r_k-r_{k-1}}$ is approaching $\infty$.
Combining the above two inequalities, we have 

\begin{equation}
\begin{aligned}
P & \left(\left(\mathcal{A}^i,\mathcal{A}^1\right)|\text{Obs}\right)  \geq \frac{1}{1 + (1-p_{q+1})^{r_k-r_{k-1}} \left( \frac{2}{|\mathcal{A}|-1}\right) ^ {r_k} + \left(\frac{1-p_{q+1}}{p_{q+1}} \cdot \frac{1}{|\mathcal{A}|-1} \right)^{r_k-r_{k-1}} },
\end{aligned}
\end{equation}
which completes our proof.

\section{Implementation Details}\label{appendix:prompt}

We use the same prompt across all datasets as in \citep{kojima2022large}. Table~\ref{tab:prompt_detail} provides examples of the prompts used for different models.

\begin{table}[htbp]
    \centering
    \begin{tabular}{p{0.15\linewidth}|p{0.75\linewidth}}
        \toprule
        QwQ & \textbf{system}: You are a helpful and harmless assistant. You are QwQ developed by Alibaba. You should think step-by-step and put your final answer within \textbackslash boxed\{\}. \newline \textbf{user}: A robe takes 2bolts of blue fiber and half that much white fiber. How many bolts in total does it take? \\
        \midrule
        Qwen3 &  \textbf{system}: You are a helpful and harmless assistant. You are Qwen developed by Alibaba. You should think step-by-step and put your final answer within \textbackslash boxed\{\}. \newline \textbf{user}: A robe takes 2bolts of blue fiber and half that much white fiber. How many bolts in total does it take? \\
        \midrule
        DeepSeek &  \textbf{system}: You are a helpful and harmless assistant. You are DeepSeek developed by DeepSeek. You should think step-by-step and put your final answer within \textbackslash boxed\{\}. \newline \textbf{user}: A robe takes 2bolts of blue fiber and half that much white fiber. How many bolts in total does it take? \\
        \bottomrule
    \end{tabular}
    \caption{Examples of prompt for different models}
    \label{tab:prompt_detail}
\end{table}

\section{Additional Results}
\label{appendix:additional_results}

\subsection{Detailed Results with Different $d_{min}$}
\label{appendix:dmin}

We report the detailed results of Table \ref{tab:sensitivity} in Table \ref{tab:dmin}. 

\begin{table}[htbp]
    % \centering
    % \small
    \setlength{\tabcolsep}{2.3pt}
    \begin{tabular}{l|cc|cc|cc|cc|cc|cc|cc}
    \toprule
     & \multicolumn{10}{c}{Math} & \multicolumn{2}{c}{Science}\\ 
     & \multicolumn{2}{c}{AIME24} & \multicolumn{2}{c}{AIME25} & \multicolumn{2}{c}{GSM8K} & \multicolumn{2}{c}{AMC} & \multicolumn{2}{c}{Olympiad} & \multicolumn{2}{c}{GPQA} & \multicolumn{2}{c}{Overall}\\ 
    Method & Acc & Tok & Acc & Tok & Acc & Tok & Acc & Tok & Acc & Tok & Acc & Tok & Acc & CR(\%) \\ 
    \midrule
    \multicolumn{15}{l}{\textit{\textbf{LLM: DeepSeek-R1-Distill-Qwen-7B}}} \\ 
    \rowcolor{gray!20} Naive & 0.60 & 11.29 & 0.37 & 11.81 & 0.91 & 1.45 & 0.88 & 6.45 & 0.57 & 8.44 & 0.45 & 9.86 & 0.63 & 100 \\ 
    $d_{min}=5$ & 0.43 & 6.79 & 0.30 & 4.71 & 0.92 & 1.13 & 0.75 & 4.18 & 0.47 & 5.10 & 0.50 & 5.01 & 0.56 & 59.00 \\ 
    $d_{min}=10$ & 0.53 & 8.15 & 0.33 & 6.97 & 0.92 & 1.16 & 0.83 & 4.72 & 0.47 & 5.69 & 0.51 & 5.68 & 0.60 & 68.23 \\ 
    $d_{min}=15$ & 0.60 & 9.45 & 0.33 & 7.65 & 0.92 & 1.18 & 0.83 & 4.88 & 0.47 & 6.20 & 0.53 & 6.07 & 0.61 & 73.42 \\
    $d_{min}=20$ & 0.60 & 9.51 & 0.33 & 8.75 & 0.91 & 1.22 & 0.83 & 5.15 & 0.48 & 6.55 & 0.53 & 6.31 & 0.61 & 77.32 \\ 
    $d_{min}=30$ & 0.60 & 10.22 & 0.33 & 9.10 & 0.91 & 1.24 & 0.88 & 5.59 & 0.48 & 6.92 & 0.52 & 7.00 & 0.62 & 82.12 \\ 
    $d_{min}=40$ & 0.60 & 11.00 & 0.33 & 9.41 & 0.91 & 1.24 & 0.88 & 5.72 & 0.48 & 7.22 & 0.50 & 7.49 & 0.62 & 85.47 \\ 
    \midrule
    \multicolumn{15}{l}{\textit{\textbf{LLM: Qwen-14B}}} \\ 
    \rowcolor{gray!20} Naive & 0.73 & 11.35 & 0.73 & 12.80 & 0.96 & 1.65 & 0.98 & 7.34 & 0.67 & 8.81 & 0.65 & 7.32 & 0.79 & 100 \\ 
    $d_{min}=5$ & 0.63 & 7.33 & 0.37 & 6.09 & 0.96 & 1.52 & 0.83 & 4.80 & 0.61 & 6.14 & 0.60 & 5.51 & 0.67 & 69.11 \\ 
    $d_{min}=10$ & 0.67 & 8.46 & 0.57 & 8.93 & 0.96 & 1.60 & 0.85 & 5.14 & 0.64 & 7.09 & 0.60 & 6.10 & 0.72 & 79.19 \\ 
    $d_{min}=15$ & 0.67 & 9.48 & 0.63 & 9.98 & 0.96 & 1.62 & 0.93 & 6.09 & 0.65 & 7.56 & 0.64 & 6.65 & 0.75 & 86.55 \\
    $d_{min}=20$ & 0.73 & 10.78 & 0.67 & 10.56 & 0.96 & 1.62 & 0.95 & 6.55 & 0.65 & 7.86 & 0.64 & 6.77 & 0.77 & 91.10 \\ 
    $d_{min}=30$ & 0.73 & 10.96 & 0.67 & 11.85 & 0.96 & 1.65 & 0.95 & 7.01 & 0.66 & 8.30 & 0.64 & 6.98 & 0.77 & 95.70 \\ 
    $d_{min}=40$ & 0.73 & 11.35 & 0.70 & 12.48 & 0.96 & 1.65 & 0.95 & 7.01 & 0.66 & 8.47 & 0.64 & 7.08 & 0.77 & 97.64 \\ 
    \midrule
    \multicolumn{15}{l}{\textit{\textbf{LLM: QwQ-32B}}} \\ 
    \rowcolor{gray!20} Naive & 0.67 & 11.14 & 0.63 & 12.03 & 0.96 & 1.43 & 0.93 & 6.80 & 0.69 & 8.16 & 0.69 & 7.45 & 0.76 & 100 \\ 
    $d_{min}=5$ & 0.57 & 5.74 & 0.40 & 6.19 & 0.96 & 1.26 & 0.90 & 4.81 & 0.58 & 4.46 & 0.61 & 4.92 & 0.67 & 63.75 \\ 
    $d_{min}=10$ & 0.63 & 7.57 & 0.50 & 7.90 & 0.96 & 1.33 & 0.90 & 5.28 & 0.61 & 5.36 & 0.63 & 5.21 & 0.71 & 73.32 \\ 
    $d_{min}=15$ & 0.67 & 8.20 & 0.57 & 9.26 & 0.96 & 1.36 & 0.93 & 5.68 & 0.64 & 6.07 & 0.65 & 5.39 & 0.74 & 79.33 \\
    $d_{min}=20$ & 0.67 & 8.77 & 0.60 & 9.77 & 0.96 & 1.39 & 0.93 & 5.98 & 0.66 & 6.48 & 0.66 & 5.63 & 0.75 & 83.34 \\ 
    $d_{min}=30$ & 0.67 & 10.68 & 0.63 & 10.45 & 0.96 & 1.41 & 0.93 & 6.21 & 0.67 & 7.07 & 0.69 & 5.98 & 0.76 & 89.93 \\ 
    $d_{min}=40$ & 0.67 & 10.90 & 0.63 & 11.69 & 0.96 & 1.41 & 0.93 & 6.21 & 0.67 & 7.43 & 0.69 & 6.22 & 0.76 & 93.25 \\  
    
    \bottomrule
    \end{tabular}
    \caption{Accuracy, number of tokens, and overall compression rate with different $d_{min}$ among 6 datasets. }
    \label{tab:dmin}
\end{table}

\subsection{Sensitivity Analysis of Different p-values. }
\label{appendix:robust}
\begin{figure}[htbp]
    \centering
    \begin{subfigure}[b]{1\linewidth}
        \centering
        \includegraphics[width=\linewidth]{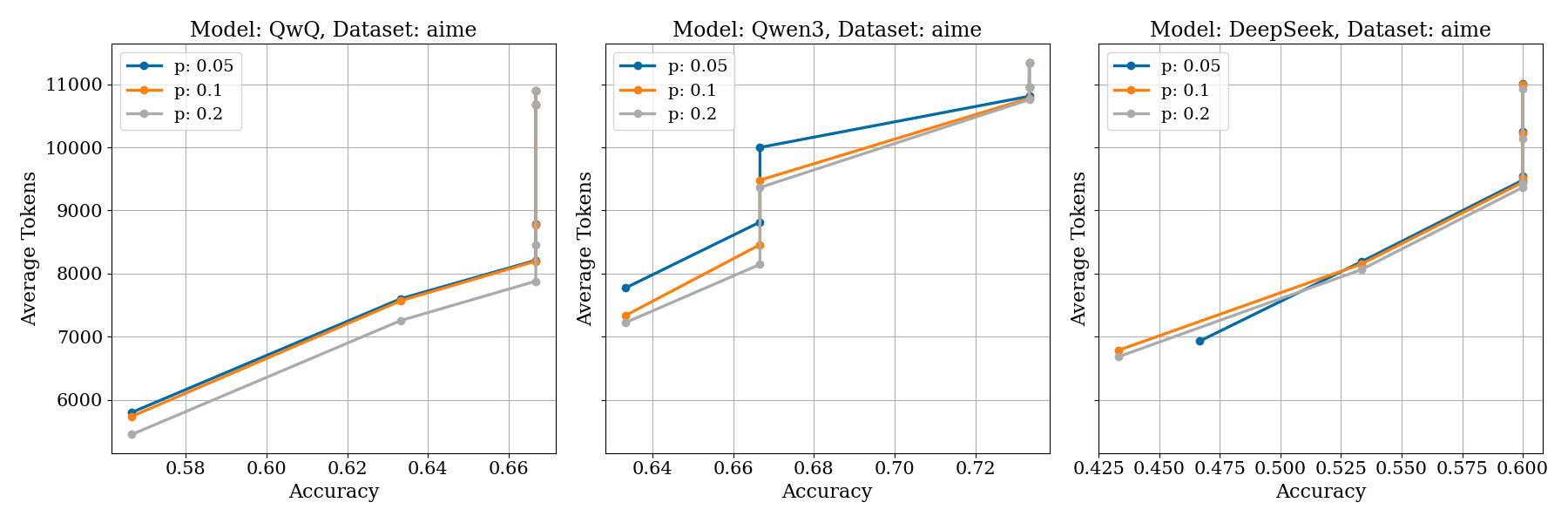}
        \caption{AIME24}
    \end{subfigure}
    \vfill
    \begin{subfigure}[b]{1\linewidth}
        \centering
        \includegraphics[width=\linewidth]{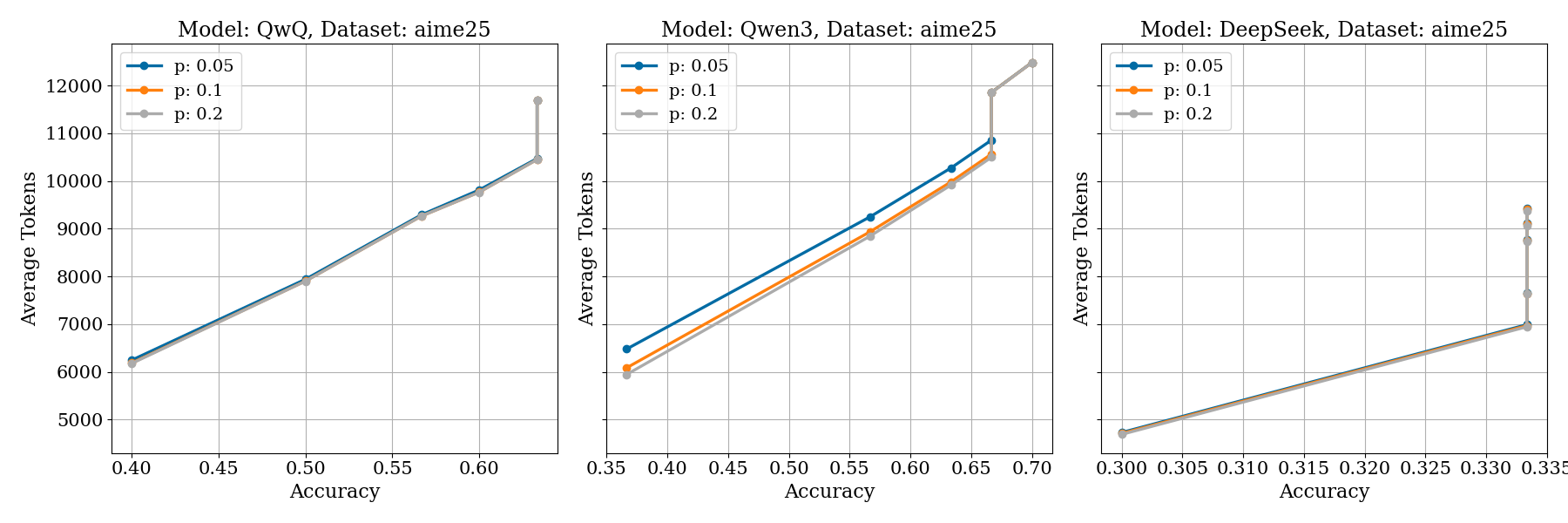}
        \caption{AIME25}
    \end{subfigure}
    \vfill
    \begin{subfigure}[b]{1\linewidth}
        \centering
        \includegraphics[width=\linewidth]{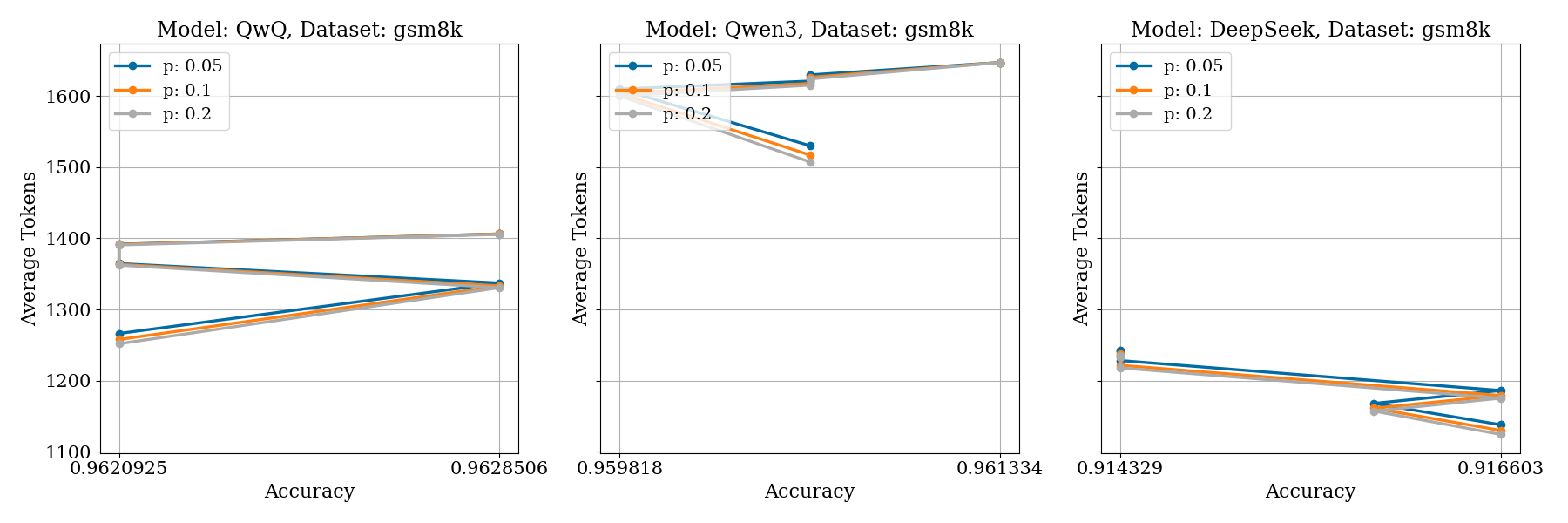}
        \caption{GSM8K}
    \end{subfigure}
    \caption{Robustness analysis of ES-CoT on the AIME24, AIME25, and GSM8K datasets regarding the hyperparameters $d_{min}$ and p-value.}
    \label{fig:pvalue1}
\end{figure}

\begin{figure}[htbp]
    \centering
    \begin{subfigure}[b]{1\linewidth}
        \centering
        \includegraphics[width=\linewidth]{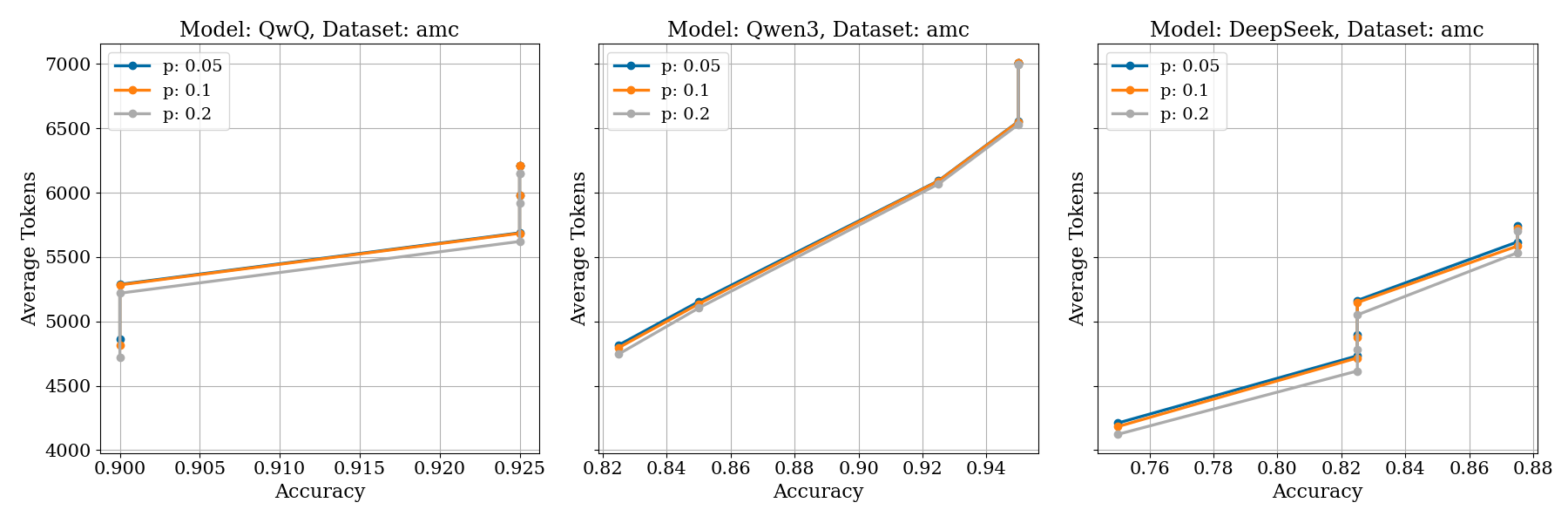}
        \caption{AMC}
    \end{subfigure}
    \vfill
    \begin{subfigure}[b]{1\linewidth}
        \centering
        \includegraphics[width=\linewidth]{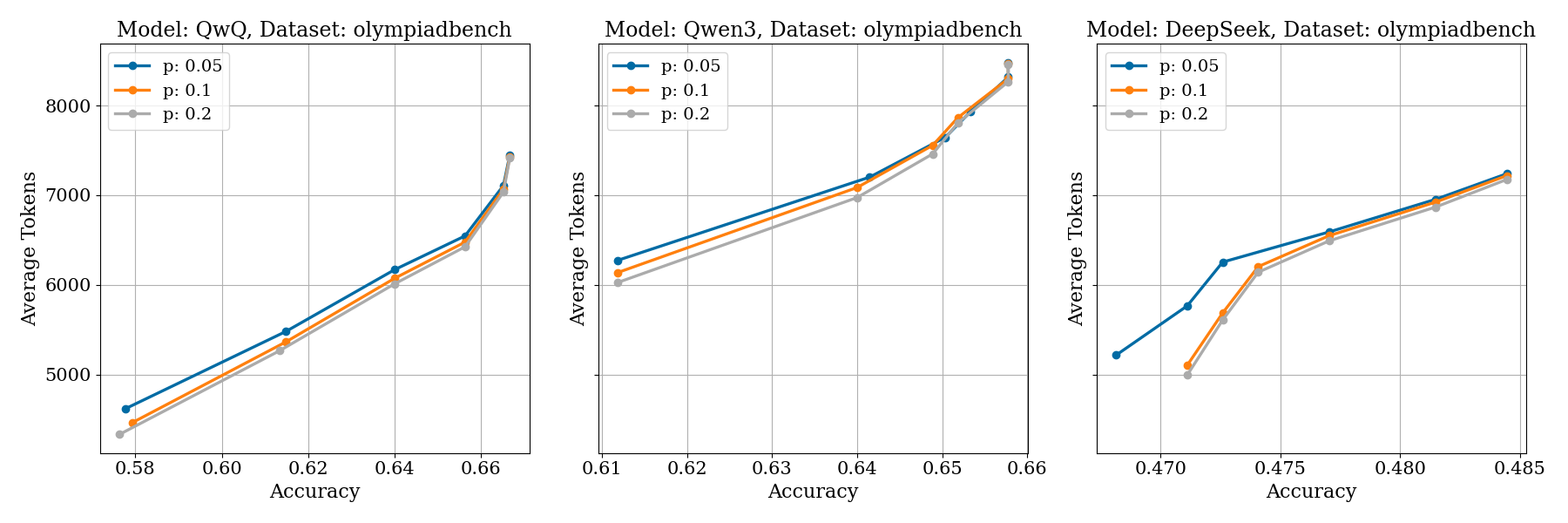}
        \caption{Olympiad}
    \end{subfigure}
    \vfill
    \begin{subfigure}[b]{1\linewidth}
        \centering
        \includegraphics[width=\linewidth]{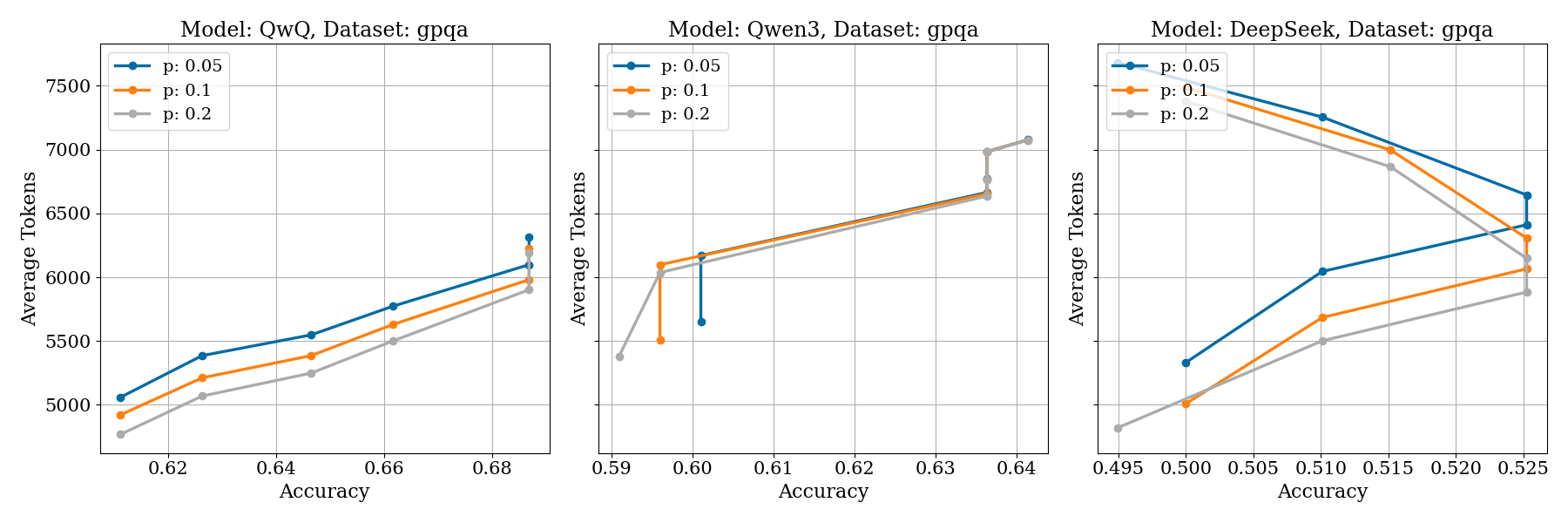}
        \caption{GPQA}
    \end{subfigure}
    \caption{Robustness analysis of ES-CoT on the AMC, Olympiad, and GPQA datasets regarding the hyperparameters $d_{min}$ and p-value.}
    \label{fig:pvalue2}
\end{figure}

% In addition to the GPQA results reported in the main text (Figure~\ref{fig:p-value}), w
We conduct the robustness analysis of the p-value threshold on all datasets and report the results in Figure~\ref{fig:pvalue1} and \ref{fig:pvalue2}. Overall, the results are consistency with previous findings in main text. ES-CoT is robust to different p-values.

\subsection{Case Study }

\begin{figure}[t]
    \centering
    \includegraphics[width=0.95\linewidth]{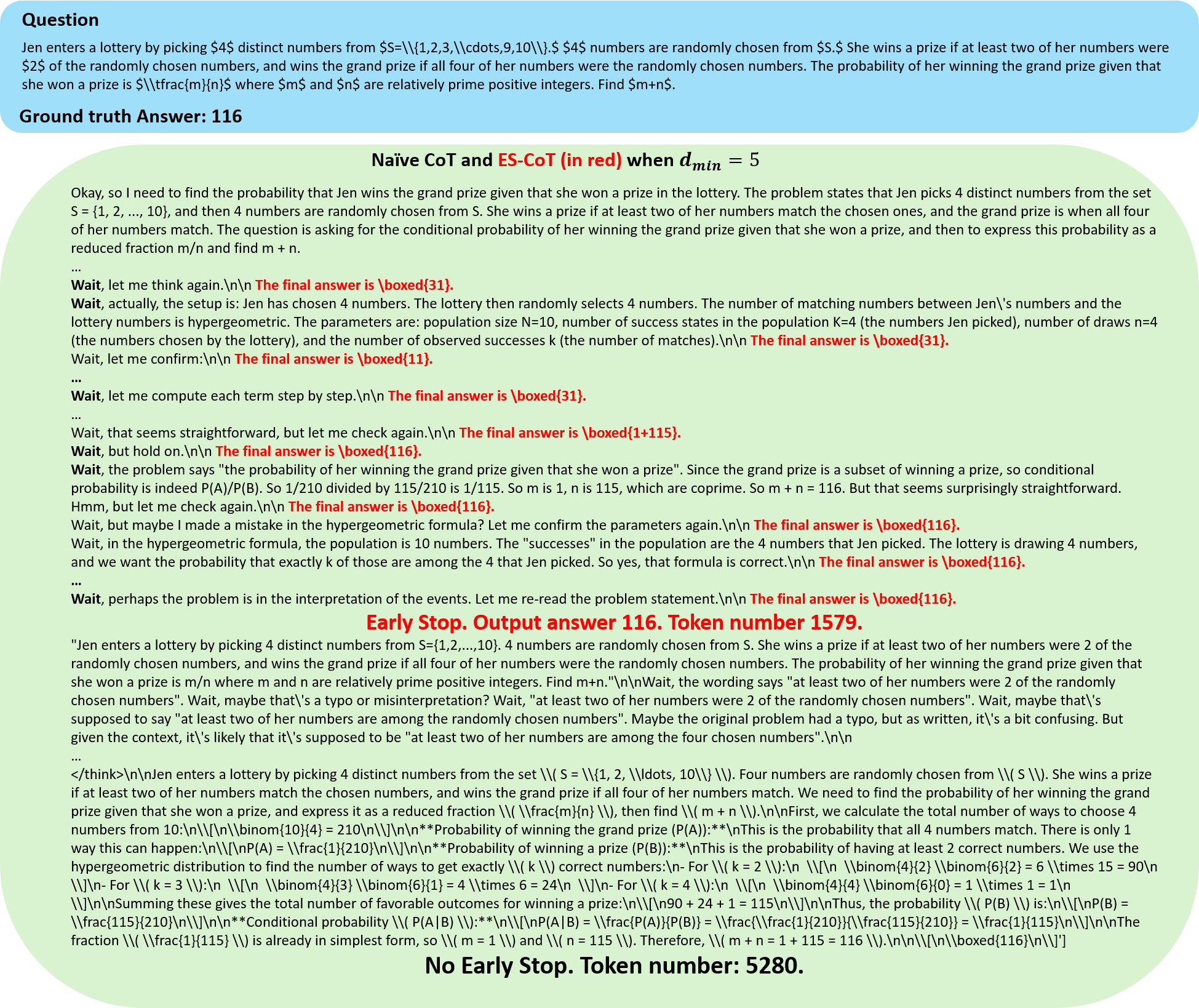}
    \caption{Comparison of generated examples by Naive and ES-CoT on AIME24.}
    \label{fig:case}
\end{figure}

We provide a case study in Figure \ref{fig:case}. This question is extracted from the AIME24 dataset with DeepSeek as the generation LLM. 

\end{document}

%% file: math_commands.tex
\usepackage{amsmath,amsfonts,bm}

\def\eqref#1{equation~\ref{#1}}
\def\1{\bm{1}}

\DeclareMathAlphabet{\mathsfit}{\encodingdefault}{\sfdefault}{m}{sl}
\SetMathAlphabet{\mathsfit}{bold}{\encodingdefault}{\sfdefault}{bx}{n}